\documentclass[conference]{IEEEtran}
\IEEEoverridecommandlockouts
\usepackage{cite}
\usepackage{amsmath,amssymb,amsfonts}
\usepackage{algorithmic}
\usepackage{graphicx}
\usepackage{textcomp}
\usepackage{xcolor}

\usepackage{multirow}
\usepackage{array}
\usepackage{threeparttable}
\usepackage{lipsum}

\usepackage{amsfonts,amssymb}
\usepackage{multirow}
\usepackage{pifont}
\usepackage{booktabs}

\usepackage[hyphens]{url}  % DO NOT CHANGE THIS

\def\BibTeX{{\rm B\kern-.05em{\sc i\kern-.025em b}\kern-.08em
    T\kern-.1667em\lower.7ex\hbox{E}\kern-.125emX}}
\begin{document}

\title{MixBoost: Improving the Robustness of Deep Neural Networks by Boosting Data Augmentation\\
%{\footnotesize \textsuperscript{*}Note: Sub-titles are not captured %in Xplore and
%should not be used}
\thanks{This paper is supported by the National Natural Science Foundation of China (Grant No. 62192783, U1811462), the Collaborative Innovation Center of Novel Software Technology and Industrialization at Nanjing University.}
}

\author{\IEEEauthorblockN{1\textsuperscript{st} Liu Zhendong}
\IEEEauthorblockA{\textit{Department of Computer Science and Technology} \\
\textit{Nanjing University}\\
Nanjing, China \\
dz20330019@smail.nju.edu.cn}\\
\IEEEauthorblockN{2\textsuperscript{nd} Wenyu Jiang}
\IEEEauthorblockA{\textit{Department of Computer Science and Technology} \\
\textit{Nanjing University}\\
Nanjing, China \\
dz21330012@smail.nju.edu.cn}\\
\and
\IEEEauthorblockN{2\textsuperscript{nd} Wenyu Jiang}
\IEEEauthorblockA{\textit{Department of Computer Science and Technology} \\
\textit{Nanjing University}\\
Nanjing, China \\
lygjwy@smail.nju.edu.cn}\\
\IEEEauthorblockN{4\textsuperscript{th} Chongjun Wang}
\IEEEauthorblockA{\textit{Department of Computer Science and Technology} \\
\textit{Nanjing University}\\
Nanjing, China \\
chjwang@nju.edu.cn}\\
}

\maketitle

\begin{abstract}
As more and more artificial intelligence (AI) technologies move from the laboratory to real-world applications, the open-set and robustness challenges brought by data from the real world have received increasing attention. Data augmentation is a widely used method to improve model performance, and some recent works have also confirmed its positive effect on the robustness of AI models. However, most of the existing data augmentation methods are heuristic, lacking the exploration of their internal mechanisms. We apply the explainable artificial intelligence (XAI) method, explore the internal mechanisms of popular data augmentation methods, analyze the relationship between game interactions and some widely used robustness metrics, and propose a new proxy for model robustness in the open-set environment. Based on the analysis of the internal mechanisms, we develop a mask-based boosting method for data augmentation that comprehensively improves several robustness measures of AI models and beats state-of-the-art data augmentation approaches. Experiments show that our method can be widely applied to many popular data augmentation methods. Different from the adversarial training, our boosting method not only significantly improves the robustness of models, but also improves the accuracy of test sets. Our code is available at \url{https://github.com/Anonymous/for/submission}.
\end{abstract}


\section{Introduction}
In recent years, the artificial intelligence (AI) models, especially deep neural networks (DNNs), have demonstrated excellent performance on popular datasets \cite{kolesnikov2020big,dosovitskiy2020image,foret2020sharpness,wortsman2022model,dai2021coatnet}. However, as AI techniques are increasingly being deployed into human society, more and more real-world data problems are raised, and AI techniques are challenged by safety and robustness issues. When deploying AI models in the real world, there are problems of robustness against distribution shifts, defense against adversarial attacks, anomaly detection, prediction consistency, calibration of prediction probabilities, etc. Data augmentation has been shown to be an effective way to overcome these challenges. For example,  \cite{yin2019fourier} 
find that some recently proposed data augmentation optimized on clean data can also improve the robustness of corrupted data. Adversarial training can also be viewed as a type of data augmentation. It improves adversarial robustness but often leads to a decrease in the accuracy performance of classification models \cite{tsipras2018robustness}. \cite{ren2021unified} explain adversarial robustness based on game interactions.
In order to establish a benchmark for evaluating out-of-distribution robustness, \cite{hendrycks2021many,hendrycks2018benchmarking} introduce some new datasets and propose data augmentation methods to improve the robustness of DNNs. With the requirements of various safety measures of AI technologies, \cite{hendrycks2022pixmix} propose a data augmentation method to achieve near Pareto-optimal of different safety and robustness metrics.

In addition to playing a role in the field of safe and robust AI, data augmentation methods are widely used in other fields, and many state-of-the-art DNNs rely heavily on data augmentation. For example, the simplest intuition-based data augmentation in computer vision tasks is to rotate, flip, or cut images. 
Although some researchers have tried to summarize data augmentation methods \cite{chen2020group,yang2022sample,mintun2021interaction,dao2019kernel}, there is still not enough understanding of them from the safety and robustness perspective. To the best of our knowledge, existing data augmentation methods used for improving safety and robustness are messy and rely heavily on intuition, without effective tools to guide algorithm design. For example, Cutout \cite{devries2017improved} simply cuts a part of the image, Mixup \cite{zhang2018mixup,tokozume2018between} uses linear interpolation to obtain new sample data, CutMix \cite{yun2019cutmix} combines Mixup and Cutout, AutoAugment \cite{cubuk2019autoaugment} is a learning-based method for data augmentation, and the idea of PixMix \cite{hendrycks2022pixmix} is to introduce new complexity into training. Although these methods have been shown to be effective in some tasks, they have still not been put together and analyzed uniformly to find their common internal mechanisms from a safety and robustness perspective.

In order to clarify the internal mechanism of existing data augmentation methods, we use XAI tools to analyze them and look for the common features of state-of-the-art data augmentation methods that positively impact the safety and robustness axis of DNNs. We also propose a method to boost data augmentation and balance these measures by controlling multi-order game interactions. Our main contributions can be summarized as follows:
\begin{itemize}
\item We analyze existing data augmentation methods that positively impact the safety and robustness measures of DNNs and summarize their common features.
\item Based on an analysis of multi-order game interaction, we design a training pipeline based on mask to control the preference of safety and robustness metrics for DNNs. This approach encourages different metrics by tuning the parameters of game interactions to meet various application scenarios.
\item Our method simultaneously boosts the state-of-the-art data augmentation methods on various safety and robustness measures by controlling the model's preference for game interactions.
\item We propose a new proxy for the global robustness metrics of DNNs based on relative game interaction strength. It's an efficient   estimation tool for
follow-up research on safe and robust AI that avoids
complex evaluations on additional datasets.

\end{itemize}

\section{Related Work}
\label{related work}
\subsection{Safety and Robustness}
The safety and robustness of AI are broad concepts, and the real-world application scenarios are very complex. To meet the requirements of safety and robustness, we usually need to consider the effects of perturbations, noise, out-of-distribution (OOD) data, attack, etc.
Several studies on the robustness of data distribution shifts have been carried out. \cite{hendrycks2018benchmarking} propose datasets with corruptions and perturbations, including CIFAR-10/100-C, ImageNet-C, ImageNet-P, etc. \cite{hendrycks2021many} introduce various artistic renditions of object classes from the original ImageNet dataset called ImageNet-R. These datasets with corruptions, perturbations and renditions have become a robustness benchmark widely used by other works \cite{tian2021geometric,herrmann2022pyramid,zhou2022understanding,bai2022improving,mao2022towards}.
\cite{mintun2021interaction} extend datasets with corruptions and introduce CIFAR-10/100-$\rm \bar C$. Calibration of prediction probabilities \cite{guo2017calibration,lakshminarayanan2017simple,hendrycks2019using} is also a challenge for AI technology. 
In the real world, we may encounter classes that are not in the training set, i.e. the OOD detection problem. \cite{hendrycks2016baseline} find that the data in the distribution generally has larger prediction probabilities than OOD data. Some works improve the anomaly detection performance of DNNs models \cite{hsu2020generalized,lee2018simple,huang2021importance}.
There are also some works on the safety of the model from the perspective of attack and defense \cite{su2019one,kurakin2018adversarial,xu2017feature}.

\subsection{Data Augmentation}
Data augmentation is a widely used method, such as the well-known heuristics in image data, including rotation, flipping, cropping, etc.
With the development of data augmentation techniques, some new methods have been proposed, such as SMOTE \cite{chawla2002smote}, SamplePairing \cite{inoue2018data}, Cutout \cite{devries2017improved}, Mixup \cite{zhang2018mixup,tokozume2018between}, CutMix \cite{yun2019cutmix}, AutoAugment \cite{cubuk2019autoaugment}, etc.
Some studies have shown that data augmentation can not only improve the prediction accuracy on clean data of DNNs, but also help the robustness of DNNs \cite{hendrycks2021many,chun2020empirical,lopes2019improving}.
PixMix \cite{hendrycks2022pixmix} based on fractal and feature visualization can not only improve the robustness of DNNs but also improve various safety measures at the same time to achieve near Pareto-optimal.

\subsection{Game Theory in Explainable Artificial Intelligence}
Game theory has been widely applied to the study of properties of DNNs. A very popular method based on Shapley values \cite{kuhn1953contributions} is SHAP \cite{lundberg2017unified} which unifies six existing XAI methods. There are some works that explain some artificial intelligence techniques from the perspective of game interaction. \cite{zhang2020interpreting} propose the multi-order interaction and try to understand and improve the dropout operation. \cite{ren2021unified} use the multi-order interaction to study the adversarial robustness of DNNs.
\cite{deng2021discovering} use the multi-order interaction to study the representation bottleneck of DNNs and observe a phenomenon that indicates a cognition gap between DNNs and human beings.

          \begin{figure*}[t]
          \includegraphics[width=\textwidth]{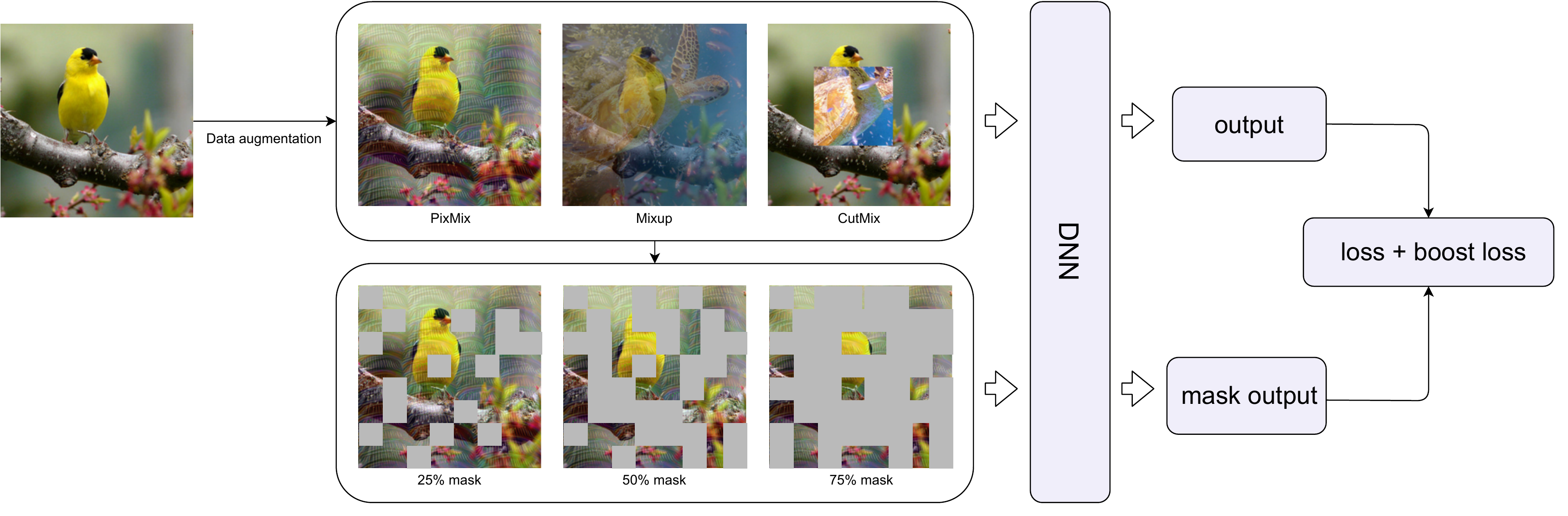}
          \caption{
          Training pipeline of the boost method of data augmentation.} \label{fig:process}
        \end{figure*}

\section{Method}
%\subsection{Feature Visualization}
%Visualization is a great way to get an intuition about how data augmentation methods work. For example, Cutout completes data augmentation by cutting out some pixels. The author believes that we can use the global information of the entire image in the neural network by using cutout, rather than the local information composed of some small features. \cite{hendrycks2022pixmix} argues that data augmentation methods need to create more complex pictures, i.e. pictures with more complex structures are more meaningful to the model. In order to intuitively understand what level the data augmentation method acts on the image dataset, we first visualized the features of various data augmentation models. Visualization methods include the popular feature importance generation methods CAM, SHAP. The models we visualize include models trained using data augmentation methods such as Cutout, Cutmix, Mixup, AutoAugment, and PixMix.
%These methods have been discussed in the related work section.

    \begin{table*}[t]
    \centering
    %\begin{tabular}{c|c|c|C{1.5cm}C{1.5cm}|C{1.5cm}C{1.5cm}|C{1.5cm}C{1.5cm}}
      \begin{tabular}{cl|c|cc|c|ccc|c}
      \hline
    \multirow{3}{*}{} & \multirow{3}{*}{} & Accuracy & \multicolumn{2}{c|}{Robustness} & Consistency &
     \multicolumn{3}{c|}{Calibration} 
    & Adversaries   \\[0.7ex]
      %\cline{3-10}
       &  & Clean & CIFAR-C & CIFAR-$\rm \bar C $ & CIFAR-P & Clean & CIFAR-C & CIFAR-$\rm \bar C $ & PGD \\
      &  & Error & mCE & mCE & mFR & RMS &RMS & RMS & Error \\
    \hline
    \multirow{8}{*}{\rotatebox{90}{CIFAR-10}} 
& Baseline &4.0 &	25.1  &	24.9  &	3.2  &6.5 &	23.9  &	23.1 &	85.7 \\
& Cutout{$\dagger$} &3.6 &	25.9 &	24.5 &	3.7 &	3.3 &	17.8 &	17.5 &	96.0 \\
&Mixup{$\dagger$} &4.2 &	21.0 &	22.1 &	2.9 &	12.5 &	12.1 &	10.9 &	93.3 \\
&CutMix{$\dagger$} &4.0 &	26.5 &	25.4 &	3.5 &	5.0 &	18.6 &	17.8 &	92.1 \\
&AutoAugment{$\dagger$} &3.9 &	22.2 &	24.4 &	3.6 &	4.0 &	14.8 &	16.6 &	95.1 \\
&AugMix{$\dagger$} &4.3 &	12.4 &	16.4 &	1.7 &	5.1 &	9.4 &	12.6 &	86.8 \\
& PixMix&3.4 &	8.0 &	11.4 &	1.7 &	3.1 &	4.0 &	4.6 &	78.7 \\
&MixBoost &\textbf{3.0} &\textbf{	6.5} &\textbf{	9.0 }&	\textbf{1.2} &\textbf{	2.3} &\textbf{	3.9} &	\textbf{4.4} &\textbf{	68.4} \\

    \hline

    \multirow{8}{*}{\rotatebox{90}{CIFAR-100}} 

&  Baseline &20.5 &	50.0 &	51.9 &	10.5 &	14.0 &	31.4 &	30.9 &	93.8  \\
& Cutout{$\dagger$}&19.9 &	51.5 &	50.2 &	11.9 &	11.4 &	31.1 &	29.4 &	98.5 \\
&Mixup {$\dagger$} &21.1 &	48.0 &	49.8 &	9.5 &	10.5 &	13.0 &	12.9 &	97.4 \\
&CutMix{$\dagger$} &20.3 &	51.5 &	49.6 &	12.0 &	12.2 &	29.3 &	26.5 &	97.0 \\
&AutoAugment{$\dagger$} &19.6 &	47.0 &	46.8 &	11.2 &	9.9 &	24.9 &	22.8 &	98.1 \\
& AugMix {$\dagger$}&20.6 &	35.4 &	41.2 &	6.5 &	12.5 &	18.8 &	22.5 &	95.6 \\
&PixMix &19.4 &	29.2 &	34.2 &	5.7 &	9.4 &	10.7 &	10.5 &	91.5 \\
& MixBoost &\textbf{18.1} &	\textbf{26.7 }&\textbf{	30.5 }	&  \textbf{ 4.7} &	\textbf{5.9} &\textbf{	7.6} &\textbf{	7.7	}& \textbf{ 86.3} \\

    \hline
  \end{tabular}
  \caption{Safety and robustness metrics of Wide ResNet trained on CIFAR-10/100 using different data augmentation methods, including clean dataset accuracy, data distribution shifts robustness, consistency, calibration of prediction probabilities, and adversarial robustness. Our MixBoost method shows the best performance on the metrics given in the table. {$\dagger$} represents results from \cite{hendrycks2022pixmix}. Lower is better for all metrics.
  }
    \label{tab1}
    \end{table*} 

\subsection{Game Interaction}

Before explaining our data augmentation boosting approach based on mask, let us first clarify some basic concepts and background. Based on the Shapley values \cite{kuhn1953contributions} in game theory, the Shapley interaction index \cite{grabisch1999axiomatic} is proposed to measure the interaction in a cooperative game.
We follow the settings of \cite{deng2021discovering} and use $v$ to denote a deep neural network that needs to be studied. In this work, we mainly study data of images, so an image containing $n$ pixels can be represented by a set of $n$ variables $N = \{1, ..., n\}$. $v(N)$ denotes the neural network output when an image containing $n$ pixels is input. The multi-order interaction $I^{(m)}(i,j)$ between two input variables $i,j \in N $ proposed by \cite{zhang2020interpreting} is defined as follows:

\begin{equation}
    I^{(m)}(i,j) = \mathbb{E}_{S\subseteq N \\{i,j}, |S|=m}
    [\Delta v(i,j,S)],\ 0 \leq m \leq n-2.
\end{equation}
where $\Delta v(i,j,S) = v(S\cup {i,j})  - v(S \cup {i}) 
  - v(S \cup {j}) + v(S)$. The $v(S)$ is the score that a DNN outputs when using variables in the subset $S\subseteq N$ only. 

\cite{deng2021discovering} find that the network outputs of a DNN can be explained as follows:
\begin{equation}
    v(N)=v(\emptyset) + \sum_{i \in N} \mu_i
    + \sum_{i,j \in N,i \neq j} \sum_{m=0}^{n-2}
    w^{(m)}I^{(m)}(i,j)
\end{equation}
where $\mu_i = v({i})- v(\emptyset)$, $ w^{(m)} = \frac{n-1-m}{n(n-1)}$.
Based on the interaction utility, \cite{deng2021discovering} find the representation bottleneck of DNNs that they tend to encode low-order and high-order interactions, and they propose a method that controls the interaction utility. Please refer to the work of \cite{deng2021discovering} for more theoretical proof details.

\subsection{Data Augmentation Boosting}

Because there is a representation bottleneck in DNNs, human beings and DNNs have different perspectives on image data. While DNNs tend to encode low-order and high-order interactions, humans may be more sensitive to mid-order interactions. Viewing pictures from a human perspective is less susceptible to perturbation and corruption, so we suggest that mid-order interactions contribute significantly to model robustness but low-order high-order interactions will do harm to model robustness.  Inspired by the representation bottleneck and the mask-based method MAE \cite{he2022masked},
we propose a training pipeline based on the mask to boost data augmentation. As shown in Figure \ref{fig:process}, we apply the data augmentation first.
We can choice any type of data augmentation, such as PixMix, CutMix, Mixup, etc. Then, we will use random patch masks for the images. The mask rate is a
hyperparameter. Then the two inputs will be sent to the DNN to get outputs. The loss function can be written as:
\begin{equation} \label{eq:loss}
   Loss(r_1,\lambda) = L(\mathbf{y},\mathbf{\hat y}) - \lambda L_{boost}(\mathbf{y},\mathbf{\hat y_{mask}}, r_1)
\end{equation}
where the $\mathbf{y}$ is the ground truth label, and $\mathbf{\hat y}$ is the output of images with no mask, $\mathbf{\hat y_{mask}}$ is the output of images with mask. the $r_1$ is mask rate, $0 \leq r_1 \leq 1$.
the item $L_{boost}$ in the loss function penalizes the DNNs to use high-order information. For example, when $r_1 = 0.8$, we will mask 80\% area of an image so that the DNNs are forced to not use high-order information but use mid-order information, just like people recognize objects in images. The $L$ is the common cross-entropy loss function used for classification. The $L_{boost}$ can be written as:
\begin{equation} 
L_{boost} = -P( \mathbf{\hat y} - \mathbf{\hat y_{mask}}) \log P( \mathbf{\hat y} -  \mathbf{\hat y_{mask}})
\end{equation}
The $\lambda$ is another parameter for aligning the two loss functions.

          \begin{figure*}[t]
          \includegraphics[width=\textwidth]{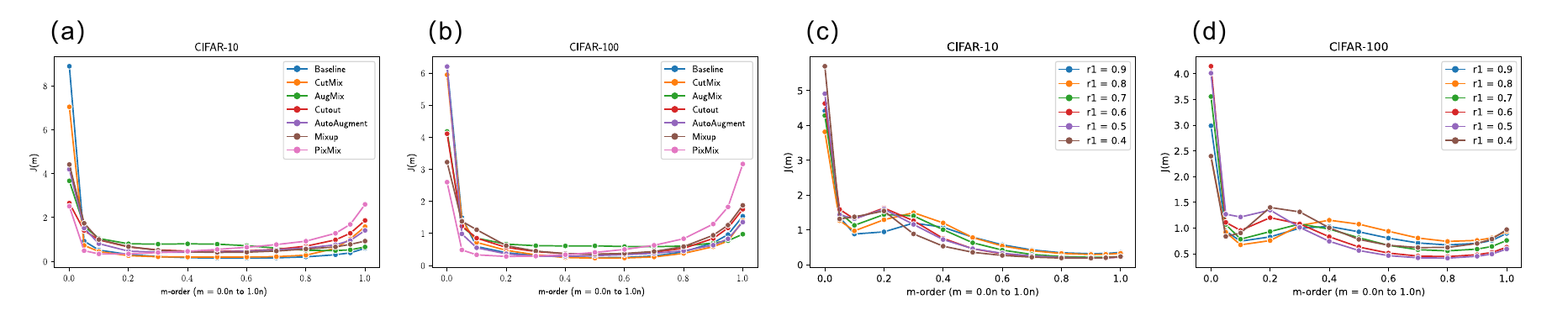}
          \caption{
          (a) and (b) are the relative game interaction strengths of Wide ResNet trained on CIFAR-10/100 using different data augmentation methods. (c) and (d) are the relative game interaction strengths of Wide ResNet training with our proposed MixBoost method on CIFAR-10/100 to penalize high-order game interactions, where $r_2$ is fixed to 1 and $\lambda_1 = \lambda_2=0 $,  $\lambda_3=1$.} \label{fig1}
        \end{figure*}
        
        \begin{figure}[t]
          \includegraphics[width=0.48\textwidth]{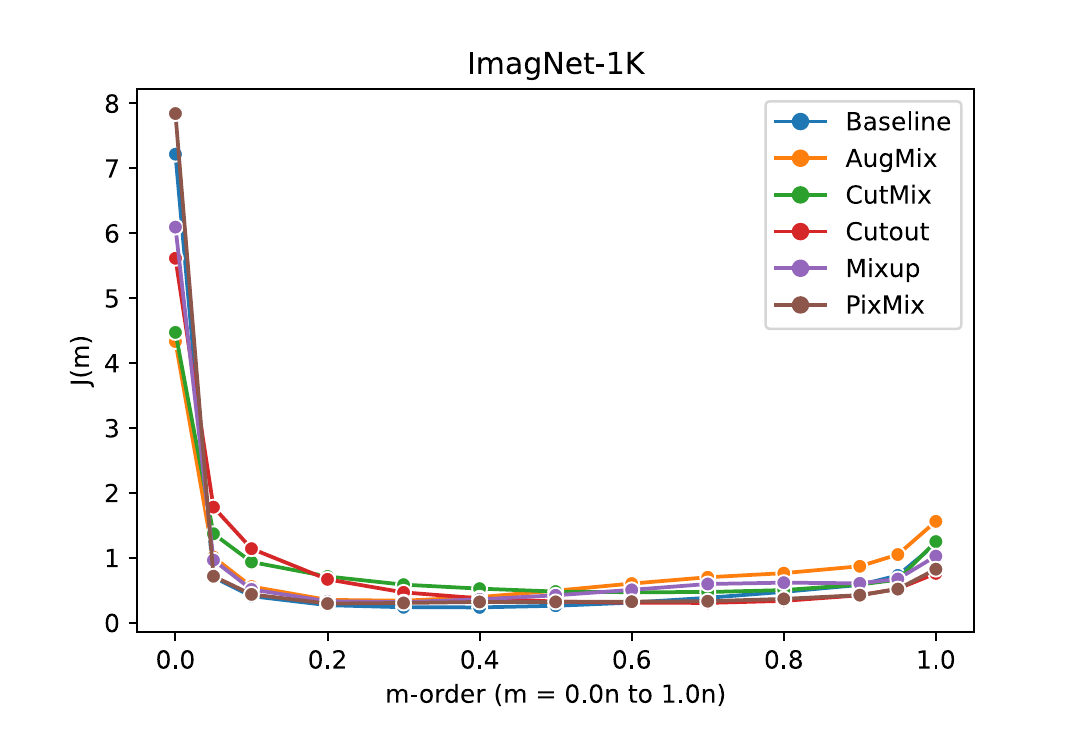}
          \caption{Relative game interaction strengths of ResNet-50 trained on ImageNet using different data augmentation methods.} 
          \label{fig:imagnet}
        \end{figure}
        
    \begin{table*}[t]
    \centering
    %\begin{tabular}{c|c|c|C{1.5cm}C{1.5cm}|C{1.5cm}C{1.5cm}|C{1.5cm}C{1.5cm}}
      \begin{tabular}{ccc|c|cc|c|ccc|c}
      \hline
    \multirow{3}{*}{} & \multirow{3}{*}{PixMix} 
    & \multirow{3}{*}{Boost} & Accuracy & \multicolumn{2}{c|}{Robustness} & Consistency &
     \multicolumn{3}{c|}{Calibration} 
    & Adversaries   \\
     %\cline{4-11}
       & & & Clean & CIFAR-C & $\rm \bar C $ & CIFAR-P & Clean &CIFAR-C & $\rm \bar C $ & PGD \\
      & & & Error & mCE & mCE & mFR & RMS &RMS & RMS & Error \\
    \hline
    \multirow{4}{*}{\rotatebox{0}{CIFAR-10}} 
& \ding{56} & \ding{56} &4.0 &	25.1  &	24.9  &	3.2  &6.5 &	23.9  &	23.1 &	85.7 \\
& \ding{52} & \ding{56} &3.4 &	8.0  &	11.6  & 1.4 &3.1  &	4.0  &	4.7  &	78.7  \\
& \ding{56} & \ding{52} &4.0 &	22.1  &	19.3  &2.6  &5.9  &	22.3 &	18.3  &	71.7  \\
& \ding{52} & \ding{52}& \textbf{3.0} &\textbf{	6.5} &\textbf{	9.0} &\textbf{	1.2} &\textbf{2.3} &\textbf{	3.9 }&\textbf{	4.4 }&	\textbf{68.4} \\

    \hline

    \multirow{4}{*}{\rotatebox{0}{CIFAR-100}} 

& \ding{56} & \ding{56} &20.5 &	50.0 &	51.9 &	10.5 &	14.0 &	31.4 &	30.9 &	93.8  \\
& \ding{52} & \ding{56}&19.4  &	29.2  &	34.2  &5.5  &	9.4  &10.7  &	10.5 &	91.5  \\
& \ding{56} & \ding{52} &21.1 &	48.3  & 46.6  & 9.7 &	11.2  &26.2  &23.5   &	87.9  \\
& \ding{52} & \ding{52} & \textbf{18.1} & \textbf{	26.7} &	 \textbf{30.5 }	&   \textbf{4.7}&	 \textbf{5.9 }& \textbf{7.6} &	 \textbf{7.7}	&  \textbf{ 86.3 }\\
    \hline
  \end{tabular}
  \caption{Results of ablation experiments to verify the effectiveness of MixBoost on CIFAR-10/100. \ding{56} means that the component is not used, and \ding{52} means that the component is used. Lower is better for all metrics.
  }
    \label{tab2}
    \end{table*} 
\subsection{A Proxy for Safety and Robustness Metrics}
We explore how safety and robustness metrics are influenced by game interactions by encouraging and penalizing specific order interactions. We do a grid search of the $\lambda$ and $r_1$ in Equation \ref{eq:loss}. Based on the mining of the relationship between the safety and robustness measures and the relative interaction strength $J^{(m)}$ defined in \cite{deng2021discovering}, we propose a proxy called adjusted mid-order relative interaction strength for the effectiveness of data augmentation on model safety and robustness improvement, defined as follows:
\begin{equation}
\label{eq:proxy}
    M(a,b,c) = \sqrt {
    \frac{1}{\max {\mathbf{J}} - \min {\mathbf{J}}}
    \frac{\sum_{m=\lfloor bn \rfloor}^{\lfloor cn \rfloor} J^{(m)}}
    {\sum_{m=0}^{\lfloor an \rfloor} J^{(m)}}
    }
\end{equation} 
where $0 \leq a \leq b \leq c \leq 1$, and $a$,$b$,$c$ are parameters that control how the order of interactions is selected.
$J^{(m)}$ is the relative interaction strength, and $n$ is the total number of variables. $\mathbf{J} = (J^{(1)},...,J^{(m)})$ is relative game interactions of all orders. The intuition of $M(a,b,c)$ can be found in Figure \ref{fig1}. Data augmentation methods that bring a more robust model tend to have lower low-order interactions and higher mid-order interactions. In fact, $M$ measures the strength of mid-order interactions relative to lower-order interactions. The method to choose $a, b, c$ is grid search, and the evaluation metric is the mean correlation between $M$ and robustness metrics.

\section{Experiments}
        
\subsubsection{Platform.}
Our experiments are mainly coded in PyTorch 1.11.0 and run on Nvidia's GPUs and Linux 5.4.0. The experiments of CIFAR-10/100 mainly run on the A5000 GPU with 24GB memory, and the experiments of ImageNet-1K (hereinafter referred to as ImageNet) mainly run on the A40 GPU with 48GB memory.

\subsubsection{Datasets.}
The training datasets we use include CIFAR-10/100 \cite{krizhevsky2009learning} and ImageNet \cite{russakovsky2015imagenet}. Both CIFAR-10 and CIFAR-100 consist of a training set of 60,000 $32\times32$ images and a validation set of 10,000 $32 \times 32$ images. ImageNet-1K is a dataset of 1000 classes with about 1.28 million images. We down-sample the images of ImageNet-1K to $224\times224$, and select 200 classes in ImageNet-R to form the ImageNet-200 dataset. To evaluate the safety and robustness of DNNs, we also use the CIFAR-10/100-C, CIFAR-10/100-$\rm \bar C$, CIFAR-10/100-P, ImageNet-C, and ImageNet-R datasets that mentioned in related work. To evaluate the model’s OOD detection performance, we use out-of-distribution datasets including Textures \cite{cimpoi2014describing}, LSUN \cite{yu2015lsun}, TinyImageNet \cite{deng2009imagenet}, and Places365 \cite{zhou2017places}. All images are down-sampled to size of $32 \times 32$.
\subsubsection{Model and Hyperparameters.}
For a fair comparison, the models we use include 40-4 Wide ResNet \cite{WideResNet} and pre-trained ResNet-50. The 40-4 Wide ResNet is used on the CIFAR10/100, and ResNet-50 is used on the ImageNet. The drop rate of the 40-4 Wide ResNet we use is 0.3 and the initial learning rate is 0.05 which follows the cosine schedule \cite{loshchilov2016sgdr}. We use the pre-trained ResNet-50 model provided by PyTorch for fine-tuning, the initial learning rate is 0.01, and the learning rate schedule is the same as Wide ResNet. For the parameters of PixMix, we follow the settings in the original paper \cite{hendrycks2022pixmix}. We search for parameters of game interaction loss function, including $\lambda_1$, $\lambda_2$, $\lambda_3$, $r_1$, and $r_2$. When we search for parameters on CIFAR-10/100, the training epochs are either 100 epochs or 200 epochs. We set training epochs to 500 when training with the searched optimal parameters. For Imagenet-1K, we set training epochs to 90. Please refer to Supplementary Materials for more details.

\subsubsection{Metrics.}
In order to measure safety and robustness of models comprehensively and fairly, 
we follow some metrics in \cite{hendrycks2022pixmix} to evaluate the model on axes of \textbf{corruptions and perturbations}, 
\textbf{consistency},
\textbf{adversaries},
\textbf{calibration},
\textbf{OOD Detection}. We use mean corruption error (mCE) to evaluate robustness against
corruptions and perturbations, and mean flip rate (mFR) to evaluate consistency. For adversarial robustness, we use projected gradient descent (PGD) \cite{madry2017towards} to generate untargeted perturbations and follow parameters in \cite{hendrycks2022pixmix}. For the task to classify images with calibrated probabilities, we use RMS calibration error \cite{nguyen2015posterior} which is defined as $\sqrt{\mathbb{E}_C [(\mathbb{P}(Y = \hat Y |C =c) - c)^2]}$, where $C$ is the confidence to predict $\hat Y$ correctly. For OOD detection, we use the area under the
receiver operating characteristic curve (AUROC) and the false positive rate (FPR) of out-of-distribution data when the true positive rate (TPR) of in-distribution data is at 95\% (FPR@95TPR).

%\subsection{Visualization of Data A  ugmentation}

\subsection{Relative Game Interaction Strength}
        
\label{ex1}

      \begin{figure}[t]
          \includegraphics[width=0.48\textwidth]{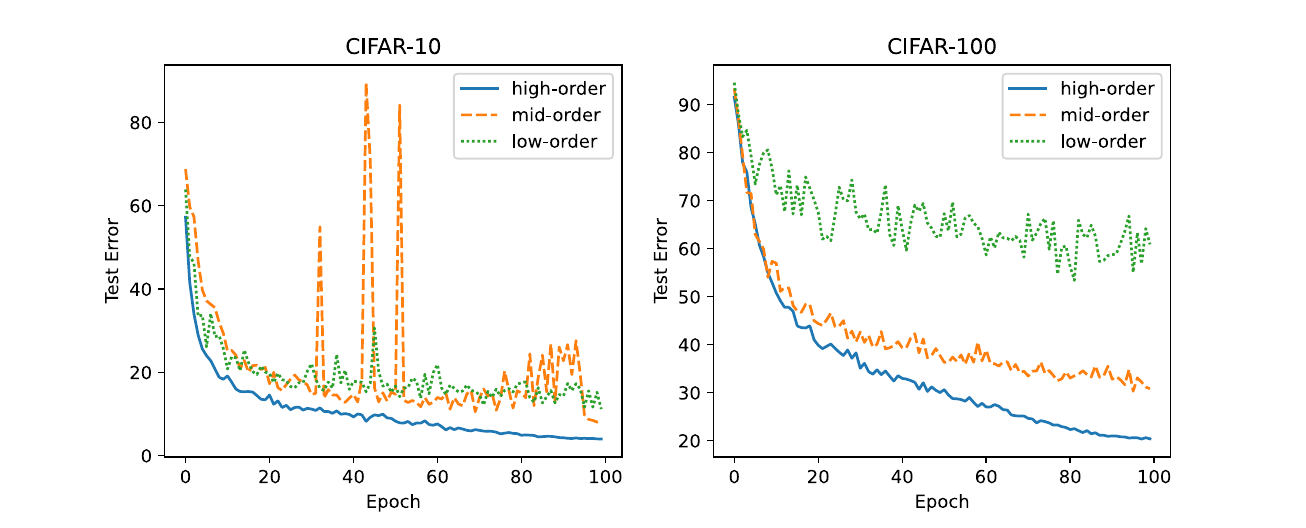}
          \caption{
          Test set error rate for Wide ResNet trained on CIFAR-10/100 for 100 epochs when penalizing high-order game interactions, penalizing low-order game interactions, and encouraging mid-order game interactions.} \label{fig2}
        \end{figure}
        
    \begin{figure*}[t]
          \includegraphics[width=\textwidth]{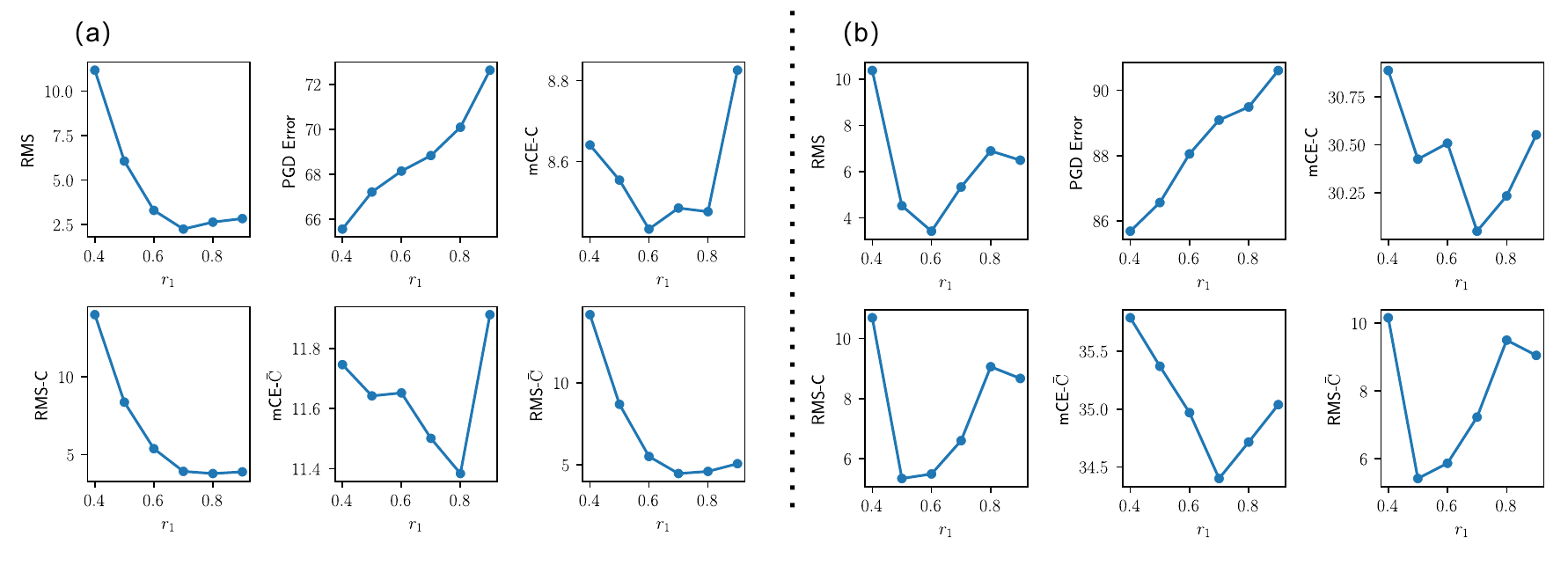}
          \caption{
Curves of some safety and robustness metrics when we fix $r_2$ to 1 and adjust $r_1$ from 0.9 to 0.4 to penalize high-order game interactions. (a) is the result of CIFAR-10 on Wide ResNet, (b) is the result of CIFAR-100 on Wide ResNet.} \label{fig5}
        \end{figure*}
        
         \begin{figure}[t]
          \includegraphics[width=0.48\textwidth]{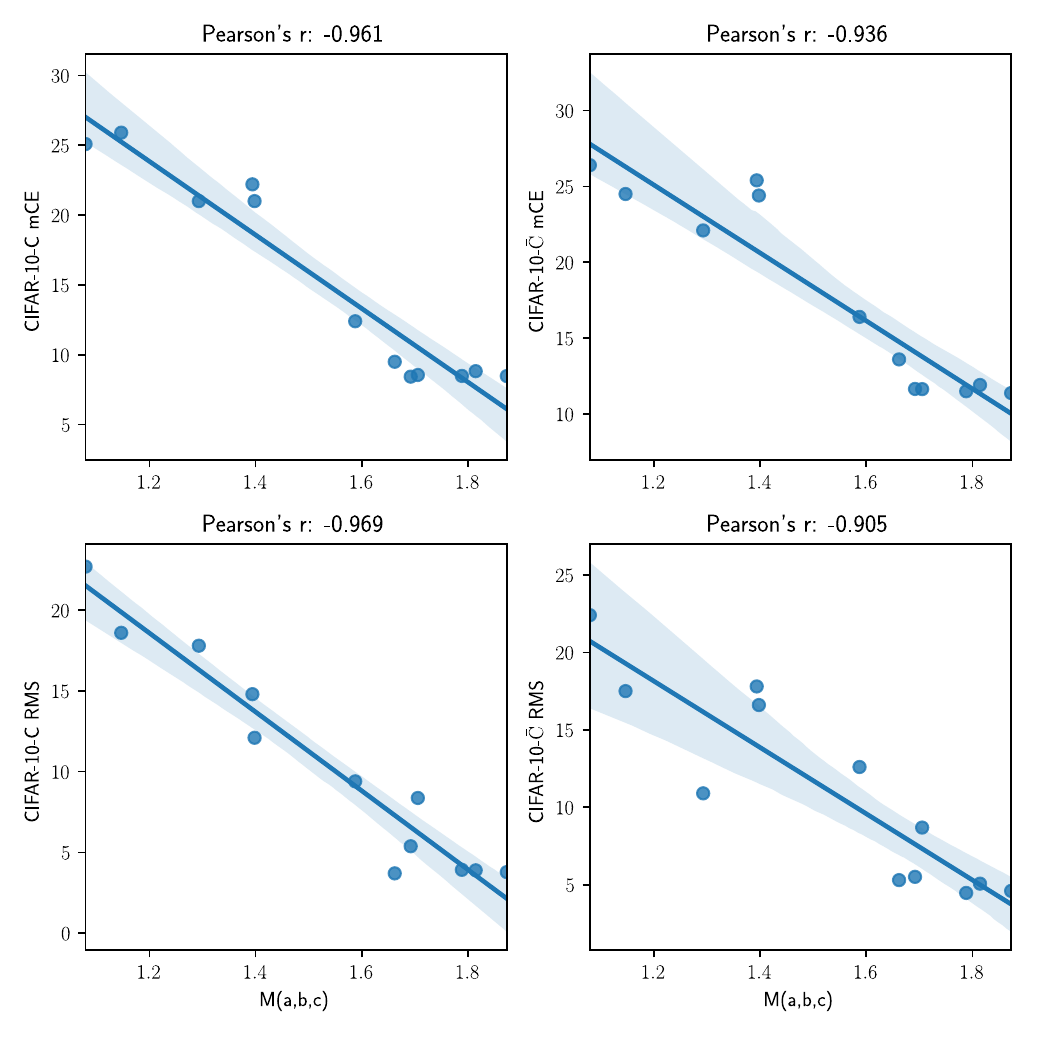}
          \caption{Relationship between the proxy $M(a,b,c)$ and some safety and robustness metrics, including robustness against corruptions and perturbations, prediction probabilities calibration when using CIFAR-10-C and CIFAR-10-$\rm \bar C$.} \label{fig3}
        \end{figure}      
        
                 \begin{figure}[t]
          \includegraphics[width=0.48\textwidth]{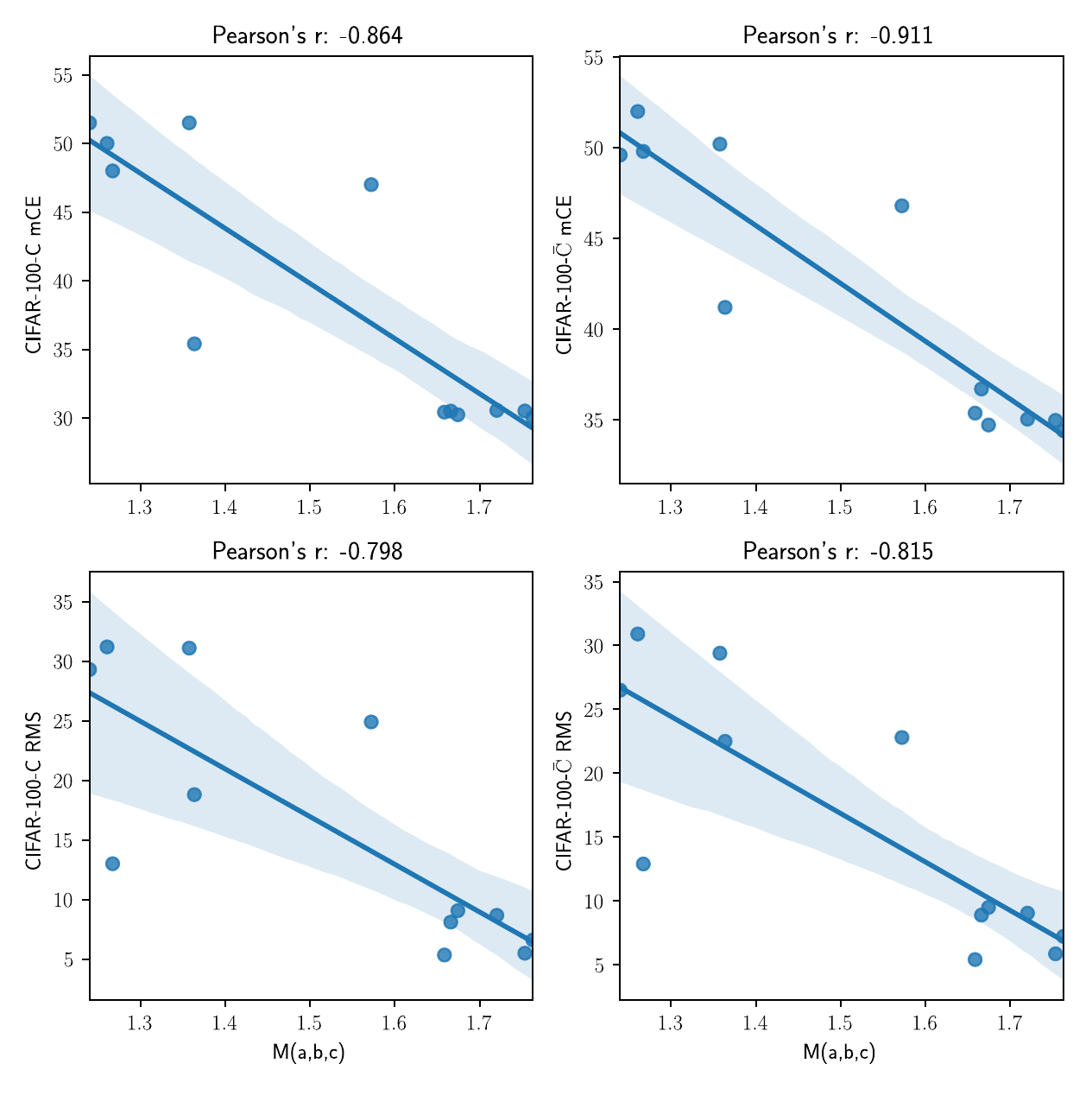}
          \caption{Relationship between the proxy $M(a,b,c)$ and some safety and robustness metrics, including robustness against corruptions and perturbations, prediction probabilities calibration when using CIFAR-100-C and CIFAR-100-$\rm \bar C$.} \label{fig4}
        \end{figure}
\subsubsection{Results of CIFAR-10/100.}

We explore some models that training with different data augmentation technologies provided by \cite{hendrycks2019augmix, hendrycks2022pixmix}. There are six data augmentation techniques: Cutout, CutMix, AutoAugment, Mixup, AugMix, and PixMix. Models trained based on these techniques outperform standard trained baseline models on safety and robustness metrics. We follow the method in \cite{deng2021discovering} to calculate the relative game interaction strength of these models, and show the results of these models in Figure \ref{fig1}. Compared with the baseline model, we find that the common feature of models training with data augmentation is to weaken low-order relative game interaction strength and encourage mid-order and high-order relative game interaction strength. This phenomenon suggests that low-order interactions contribute less to the model's safety and robustness than mid-order and high-order interactions. Our experiments also validate the conclusion in \cite{deng2021discovering} that mid-order interactions are hard to learn for DNNs without additional constraints. The best method for balancing various safety and robustness metrics is PixMix, and it has the smallest low-order interactions and the highest high-order interactions. This inspires us that if we want to improve the safety and robustness of a deep learning model, we need to \textbf{pay more attention to the mid-order and high-order interactions}, and have \textbf{a global understanding like human cognition rather than limited to small local information}.
        
\subsubsection{Results of ImageNet.} We also explore the relative game interaction strength of some ResNet-50 models trained using data augmentation techniques, including PixMix, AugMix, Cutout, Mixup, and CutMix. The results are shown in Figure \ref{fig:imagnet}. The results on ImageNet are similar to CIFAR-10/100, but it is clear that high-order interactions are not well modeled, and most models only model low-order interactions. Knowledge representations for safety and robustness are not well modeled, except for AugMix. Experiments using game interaction also demonstrate that ResNet-50 with PixMix does not model the robustness problems of the ImageNet dataset well, which may be the reason for the poor performance.
Please refer to the Supplementary Materials for more details.

\subsection{Boosting Data Augmentation.}
        
\subsubsection{Safety and Robustness Metrics.}
        
\label{ex2}
Inspired by Figure \ref{fig1}(a) and Figure \ref{fig1}(b), we conjecture that controlling the relative game interaction strength can effectively improve the safety and robustness of DNNs. We verify this conjecture and improve the safety and robustness of DNNs. Table \ref{tab1} shows the results of CIFAR-10/100. When discussing safety and robustness metrics, we have chosen the average of 5 runs. We also use the Wilcoxon signed-rank test, setting the level of significance to 0.05 to test the significance of performance improvement.
%and Table \ref{tabx} shows results of ImageNet-200. 
Our experiments are based on the state-of-the-art data augmentation method PixMix. We comprehensively improve the robustness metrics by using our training pipeline. We call this approach MixBoost. We omit the details of parameter searching and present only the results with best parameters $r1 = 0.7$. Please refer to the Supplementary Materials for more details.

The results in Table \ref{tab1} show that on the CIFAR-10/100, MixBoost not only improves the classification accuracy on clean data but also outperforms PixMix on various safety and robustness metrics. For example, the classification accuracy of the corrupted dataset CIFAR10-C is improved from 8.0\% to 6.5\%, and the robustness against attacks is improved from 78.7\% to 68.4\%.  Similarly, MixBoost improves the safety and robustness of DNNs on CIFAR-100. Our method outperforms PixMix by more than 10\% on most metrics. Please refer to the Supplementary Materials for the results of anomaly detection.
%Table \ref{tabx} shows that on the imagenet200 (200-class) dataset, MixBoost also helps......TODO TODO TODO

   % \begin{table}[t]
  %  \centering
    %\begin{tabular}{c|c|c|C{1.5cm}C{1.5cm}|C{1.5cm}C{1.5cm}|C{1.5cm}C{1.5cm}}
   %   \begin{tabular}{c|c|cc|cc}
   %   \hline
  %  \multirow{3}{*}{}  & Accuracy & \multicolumn{2}{c|}{Robustness} & \multicolumn{2}{c}{Calibration}  \\
     %\cline{4-11}
  %     &  Clean & C  & R & Clean & C \\
  %    &  Error & mCE & mCE & RMS &RMS  \\
  %  \hline

%Baseline  &4.0 &	25.1  &	24.9  &	3.2  &6.5 \\
%Baseline  &4.0 &	25.1  &	24.9  &	3.2  &6.5 \\
%Baseline  &4.0 &	25.1  &	24.9  &	3.2  &6.5 \\

 %   \hline
 % \end{tabular}
 % \caption{TODOTODOTODOTO
%  }
 %   \label{tabx}
 %   \end{table} 
    
In order to further verify the effectiveness of MixBoost, we conduct ablation experiments, and the results are shown in Table \ref{tab2}. The results in Table \ref{tab2} show that using the method of controlling game interaction alone can effectively improve the prediction probabilities calibration metric, but it will bring performance loss. Using PixMix alone is better than the baseline, but using it with our boost method will achieve better performance. This shows that both our Boost method and PixMix components are indispensable.
\subsubsection{More data Augmentation Methods with Boost}
We also use other data augmentation methods for comparison, the results can be found in \ref{tab:moreaug}. Our method can be applied to other data augmentation methods without any changes in details. We have done experiments on other advanced methods including AutoAugment \cite{cubuk2019autoaugment}, AugMix \cite{hendrycks2019augmix}, TrivialAugmentWide \cite{TrivialAugment}. The results show that most robustness metrics are improved by more than 10\% when using our boost method.

    \begin{table}[t]
    \centering
    %\begin{tabular}{c|c|c|C{1.5cm}C{1.5cm}|C{1.5cm}C{1.5cm}|C{1.5cm}C{1.5cm}}
      \begin{tabular}{c|c|c|cc|c}
      \hline
 \multirow{2}{*}{}
       &  Clean & Corrupt & Clean & Corrupt  & PGD \\
      & Error & mCE  & RMS &RMS & Error \\
    \hline

 AutoAug& 4.36 & 16.06& 4.19& 12.24& 88.91\\
+Boost& 4.35 & 14.87& 3.04&  9.77& 80.50\\

    \hline

 AugMix& 4.35 & 12.59& 5.45&  10.16& 85.42\\
+Boost& 4.01 & 10.86& 3.96&  9.04& 77.87\\

    \hline
     Trivial & 3.64  & 14.53& 2.79& 9.78& 89.52\\
+Boost& 3.67 & 13.82&  2.44& 7.42& 80.96\\

    \hline
  \end{tabular}
  \caption{Experiments on other 3 data augmentation methods including AutoAugment \cite{cubuk2019autoaugment}, AugMix \cite{hendrycks2019augmix}, TrivialAugmentWide \cite{TrivialAugment} for verifing The effectiveness of our boost method on CIFAR-10. 
  }
    \label{tab:moreaug}
    \end{table} 
    
\subsubsection{Game Interaction.}
We search the parameters of the loss function in Equation \ref{eq:loss}, some of the results are shown in Figure \ref{fig1}(c), Figure \ref{fig1}(d). Please refer to the Supplementary Materials for more details. 
We calculate the relative game interaction strength of DNNs after applying our boost method, and find an interesting phenomenon in the experiment of penalizing high-order interactions, which we call it \textbf{Long-Rope Effect}, as shown in Figure \ref{fig1}(c) and Figure \ref{fig1}(d): when you When shaking one end of the rope, the undulations of the rope are transmitted forward instead of being held in place. In our experiments, \textbf{when we penalize the high-order interactions, the mid-order interactions are boosted}, which looks a lot like what happens when we shake the rope. Penalizing high-order interactions not only suppresses high-order interactions, but can also significantly influence mid-order interactions, as well as low-order interactions. This phenomenon shows that when the model learns knowledge representation if we penalize the high-order interaction, the model will tend to learn mid-level information in order to obtain enough information, which is a behavior that conforms to the characteristics of human cognition. Humans can determine which class an image belongs to by looking at only a portion of the image without looking at the entire image. Therefore, by penalizing high-order interactions when training with data augmentation, \textbf{the model breaks through the representation bottleneck and learns more mid-order features}, which may be the reason for the improved robustness. Our penalty on high-order interactions can effectively improve the robustness and safety metrics of the model. Penalizing the high-order interaction when using data augmentation is a better way to model robustness problems.

In addition to this, we also find that when penalizing low-order interactions or mid-order interactions alone, the model does not converge well, and the training process is shown in Figure \ref{fig2}. The model fails to converge to the same level as training of penalizing the high-order interactions, which suggests that when penalizing low-order or mid-order interactions alone, the model does not model the robustness problems well, the data augmentation and interaction constraints conflict. This indicates that the \textbf{low-order and mid-order interactions may contain the necessary information for model learning}, and penalizing the necessary information will adversely affect the problem modeling. We need to take advantage of the long-rope effect to encourage mid-order interactions by penalizing high-order interactions.

\subsubsection{Relationship between Game Interaction and Safety metrics.}
Through experiments of parameters searching, we find that only penalizing high-order interactions is consistent with modeling safety and robustness problems. Therefore, we set the training epochs to 100 to study the effect of the value of $r_1$ on the robustness and safety of the model. The results are shown in Figure \ref{fig5}. We find that the safety and robustness metrics on CIFAR-10/100 exhibit similar regularities when the range of penalizing of high-order interactions expands. Except that the adversarial attack robustness decreases as $r_1$ decreases, the data distribution shifts robustness and calibration performance metrics both decrease first and then increase. These metrics are influenced by $r_1$ so we can balance them by adjusting the scope and strength of game interactions. For tasks that only have a single safety metric requirement, such as robustness against distribution shifts, robustness against perturbations, and adversarial robustness, we can adjust the $r_1$ to meet a single task metric requirement while ignoring other safety metrics.

\subsection{A Proxy for Safety and Robustness}
Our results demonstrate that penalizing high-order interactions and encouraging mid-order interactions are beneficial for safe and robust DNNs. Too many low-order interactions are detrimental to safe and robust DNNs, but low-order interactions contain information that is indispensable for DNNs. Therefore, we use the proxy $M(a,b,c)$, where $a=b=0.2$, $c=0.8$. The relationship between our proxy and some safety and robustness metrics is shown in Figure \ref{fig3} and Figure \ref{fig4}. For more results, please refer to Supplementary Materials. The Pearson correlation coefficient values of our $M(a, b, c)$ and various metrics are almost all above 0.9 in CIFAR-10 and about 0.8 in CIFAR-100, indicating that the linear correlation is significant. There is a significant linear correlation between the proxy we propose and a variety of popular safety and robustness measures within a certain range. Our proxy $M(a, b, c)$ can be used to estimate the safety and robustness without calculating many complex metrics of a model, and it's an efficient global estimation tool for follow-up research on safe and robust AI. When using our proposed agent, only the relative game interaction strength needs to be calculated without using additional datasets. Our proxy avoids the complexity of using various additional datasets and their biases.

\section{Conclusion}
We explore the internal mechanisms of various popular data augmentation methods from the perspective of safety and robustness and find that the common feature of them is to encourage the strength of mid-order relative game interactions. Based on our discovery of the long-rope effect, we propose a mask-based boosting method for data augmentation, MixBoost, which simultaneously achieves state-of-the-art performance on various robustness and safety tasks. The single requirement for different robustness and safety metrics in different application scenarios can also be met by tuning the parameters of our method. We also propose a global estimation proxy for the safety and robustness of DNNs, which is a simple yet effective tool.

\bibliographystyle{IEEEtran}
\bibliography{IEEEabrv,mybibliography}

\end{document}